\documentclass{article}
\usepackage{spconf,amsmath,graphicx}
\usepackage{amsthm}

\usepackage{amsmath}
\usepackage{dsfont}
\usepackage{algorithm}
\usepackage{mathtools}
\usepackage{algpseudocode}
\usepackage{algpseudocode}
\usepackage{amsfonts}
\usepackage[bottom]{footmisc}
\usepackage[accsupp]{axessibility}  
\usepackage{subcaption}

\usepackage[pagebackref,breaklinks,colorlinks]{hyperref}
\usepackage{url}

\newcommand\scalemath[2]{\scalebox{#1}{\mbox{\ensuremath{\displaystyle #2}}}}

\title{ProbMCL: Simple Probabilistic Contrastive Learning for Multi-label Visual Classification}
\name{Ahmad Sajedi$^{1, 2}$, Samir Khaki$^{1}$, Yuri A. Lawryshyn$^{2}$, Konstantinos N. Plataniotis$^{1}$}
\address{$^{1}$ The Edward S. Rogers Sr. Department of Electrical \& Computer Engineering, University of Toronto\\
$^{2}$ Centre for Management of Technology \& Entrepreneurship (CMTE), University of Toronto \\
\tt\small Webpage: \href{https://ahmadsajedii.github.io/ProbMCL/}{https://github.com/DataDistillation/DataDAM}}

\begin{document}
\ninept
\maketitle
\begin{abstract}
Multi-label image classification presents a challenging task in many domains, including computer vision and medical imaging. Recent advancements have introduced graph-based and transformer-based methods to improve performance and capture label dependencies. However, these methods often include complex modules that entail heavy computation and lack interpretability. In this paper, we propose Probabilistic Multi-label Contrastive Learning (ProbMCL), a novel framework to address these challenges in multi-label image classification tasks. Our simple yet effective approach employs supervised contrastive learning, in which samples that share enough labels with an anchor image based on a decision threshold are introduced as a positive set. This structure captures label dependencies by pulling positive pair embeddings together and pushing away negative samples that fall below the threshold. We enhance representation learning by incorporating a mixture density network into contrastive learning and generating Gaussian mixture distributions to explore the epistemic uncertainty of the feature encoder. We validate the effectiveness of our framework through experimentation with datasets from the computer vision and medical imaging domains. Our method outperforms the existing state-of-the-art methods while achieving a low computational footprint on both datasets. Visualization analyses also demonstrate that ProbMCL-learned classifiers maintain a meaningful semantic topology.
\end{abstract}
\section{Introduction} \label{sec:intro}
Convolutional neural networks (CNNs) excel at various computer vision tasks, particularly image classification \cite{he2016deep, sajedi2023datadam, tan2019efficientnet, sajedi2022subclass, Khaki_2023_CVPR, amer2021high, sajedi2021efficiency}. Despite numerous proposed architectures for single-label classification \cite{he2016deep, tan2019efficientnet}, progress in multi-label classification has been limited. Multi-label image classification is a fundamental task in the computer vision \cite{lin2014microsoft, chen2019multi, zhu2021residual} and medical imaging \cite{hosseini2019atlas, sajedi2023end} domains, which assigns multiple labels to an input image. Its core challenge lies in establishing embedded representations to capture meaningful labels and their interdependencies, thereby enhancing classification performance. Recent methods such as RNN-CNN for sequence-to-sequence modeling \cite{wang2016cnn}, graph-based approaches \cite{chen2019multi, wang2020multi}, transformer-based learning \cite{zhu2021residual, zhao2021transformer}, and new loss function designs \cite{ridnik2021asymmetric, sajedi2023new} have aimed to effectively learn these representations. However, these approaches fail to capture label correlations fully \cite{sajedi2023end, ridnik2021asymmetric} or impose constraints on the learning process by incorporating resource-intensive components like correlation matrices, graph neural networks, and spatial attention modules \cite{chen2019multi, zhao2021transformer}. These limitations prevent scalability to large models or datasets and hinder the execution, optimization, or interpretation of such approaches \cite{zhu2021residual}.

Contrastive learning, introduced in \cite{sajedi2023end, chen2020simple, khosla2020supervised}, aims to create useful latent embeddings by positioning similar (or positive) sample representations close and dissimilar (or negative) ones far apart. Therefore, the choice of positive and negative samples for an anchor is crucial for achieving good performance. For example, in self-supervised contrastive learning \cite{chen2020simple}, positive samples come from anchor-augmented images, while negative samples include all the other images in the batch. In supervised contrastive learning \cite{khosla2020supervised}, same-label images of the anchor are treated as positives and different-label images as negatives. However, these methods were designed for single-label tasks and cannot consider label relationships in multi-label scenarios. Adapting them to situations with multiple labels poses the challenge of determining which samples should be considered positive or negative for anchors with more than one label. Multi-label images lack the simplicity of single-label tasks, which rely on a single object for alignment. Furthermore, these latent spaces lack the necessary smoothness and structures to discern the encoder network's epistemic uncertainty \cite{chen2020simple, khosla2020supervised}. This deficiency impacts confidence levels in robust multi-label classification tasks and introduces errors in interpreting output predictions. As a result, our motivation stems from the need to capture label dependencies and investigate encoder uncertainty in multi-label classification tasks while reducing training costs.

In this paper, we propose a new framework called ``{Prob}abilistic {M}ulti-label {C}ontrastive {L}earning ({\texttt{ProbMCL}})" to leverage simple yet effective representation learning for multi-label image classification. The proposed framework introduces positive samples that share labels with the anchor image at a specific decision threshold, based on the predefined overlapping index function. Unlike existing multi-label methods \cite{chen2019multi, zhu2021residual, wang2020multi, zhao2021transformer}, \texttt{ProbMCL} avoids reliance on heavy-duty label correlation modules. Specifically, our method captures label dependencies by pulling together the feature embeddings of positive pairs and pushing away negative samples that do not share classes beyond the decision threshold. In the learning process, multi-label feature embeddings are represented using Gaussian Mixture Models (GMMs) in the probability space \cite{ sajedi2023new, varamesh2020mixture}, and probabilistic contrastive learning is used within this space to capture label correlations and the uncertainty associated with the model. The main contributions of our study include:
\begin{figure*} 
    \centering
    \includegraphics[width= 1.0\linewidth]{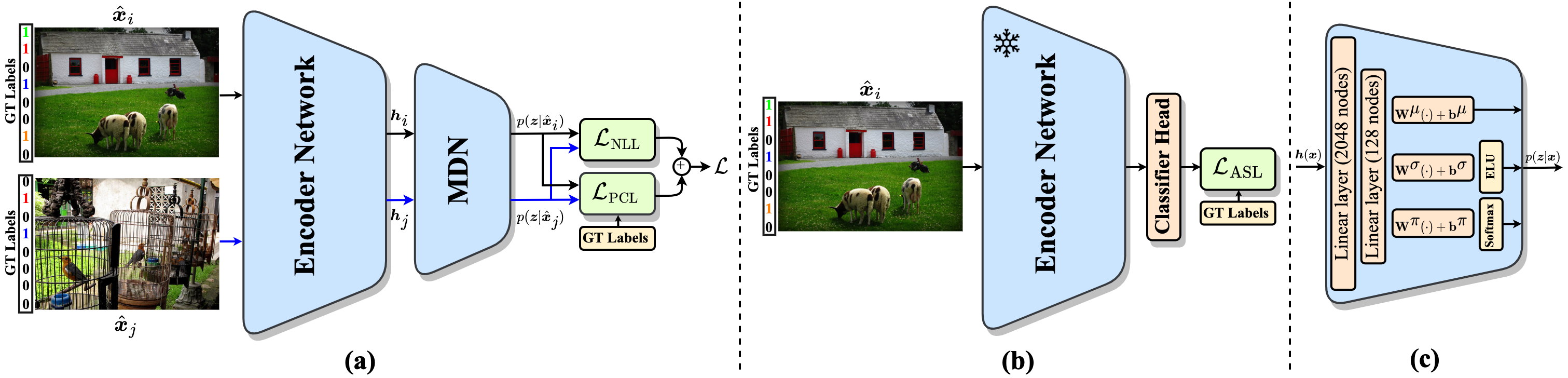}
    \caption{Illustration of the ProbMCL framework in (a) the contrastive stage and (b) the classification stage. In the classification stage, the MDN is discarded and the trained encoder is retained. (c) The internal architecture of the Mixture Density Network (MDN).}
    \label{fig:probmcl}
\end{figure*}

\textbf{[C1]:} We propose a framework that applies contrastive learning to multi-label image classification to efficiently capture label dependencies. Our loss function allows for the removal of heavy-duty label correlation modules while achieving optimal performance.

\textbf{[C2]:} We integrate a mixture density network into contrastive learning to generate mixtures of Gaussian and enhance representation learning by estimating feature encoder epistemic uncertainty.

\textbf{[C3]:} We employ our pipeline in the computer vision and computational pathology domains to showcase its effectiveness for multi-label image classification across different applications.

\section{Methodology} \label{sec:method}
In this section, we introduce a simple framework for multi-label visual representations, termed \textbf{Prob}abilitic \textbf{M}ulti-label \textbf{C}ontrastive \textbf{L}earning (\texttt{ProbMCL}). Specifically, for each input sample in a batch, we generate two distinct augmented versions using the RandAugment technique \textit{Aug}($\cdot$) from \cite{sajedi2023end, wang2020multi}. These augmentations provide unique perspectives on the original sample, capturing specific subsets of its information while guaranteeing label preservation in augmented images. An encoder network then processes the augmented samples to derive feature embeddings. During training, the feature embeddings are transformed using a mixture density network (MDN) to generate Gaussian mixture model (GMM) parameters. This GMM produces the output feature's probability density function in a space where probabilistic contrastive learning is defined. Positive samples are carefully chosen based on similarity measures and the overlap decision threshold. In inference, the MDN is discarded, and the trained encoder \textit{Enc}($\cdot$) is retained. A classification head is added to the feature encoder and trained on a multi-label classification task with a frozen \textit{Enc}($\cdot$) using an asymmetric loss function \cite{ridnik2021asymmetric}. Figure \ref{fig:probmcl} provides a visual explanation of the proposed \texttt{ProbMCL} framework.

\subsection{ProbMCL Framework} \label{subsec:probmcl}
The main components of our framework are:

\textbf{Encoder Network.} The encoder network, represented as \textit{Enc}($\cdot$), derives feature representation vectors $\boldsymbol{h}(\boldsymbol{x}) \in \mathbb{R}^{H}$ from augmented data $\boldsymbol{x}$. Aligned with state-of-the-art practices \cite{khosla2020supervised, wang2020understanding}, we normalize these vectors onto the unit hypersphere within $\mathbb{R}^{H}$. The \texttt{ProbMCL} framework affords flexibility in selecting network architectures without imposing any limitations.

\textbf{Mixture Density Network.} 
The Mixture Density Network (MDN) transforms encoder-derived representation vectors into parameters of a Gaussian mixture, including mixture coefficients $\pi_{k}$, mean vector $\boldsymbol{\mu}_{k}$, and covariance matrix $\boldsymbol{\Sigma}_{k}$ for each $k$th component. This generates a probability density function for the output feature $\boldsymbol{z} \in \mathbb{R}^{n}$ conditioned on $\boldsymbol{x}$, where the contrastive loss is applied. Indeed, the MDN operates as a Gaussian mixture model, defined as:
\vspace{-11pt}
\begin{flalign} \label{eq:gmm}
   p(\boldsymbol{z}|\boldsymbol{x}) = \sum_{k=1}^{C} \pi_{k}(\boldsymbol{x}) \mathcal{N}\Big(\boldsymbol{z}; \mu_{k}(\boldsymbol{x})\boldsymbol{1}, \sigma^{2}_{k}(\boldsymbol{x})\boldsymbol{I}\Big),
\end{flalign}
where $\sum_{k=1}^{C} \pi_{k}(\boldsymbol{x}) = 1$, $\pi_{k}(\boldsymbol{x}) \geq 0$, $\forall k \in \{1, \cdots, C\}$ and $\mathcal{N}\big(\boldsymbol{z}; \mu_{k}(\boldsymbol{x})\boldsymbol{1}, \sigma^{2}_{k}(\boldsymbol{x})\boldsymbol{I}\big)$ is an isotropic normal distribution with a mean vector $\boldsymbol{\mu}_{k} = \mu_{k}(\boldsymbol{x})\boldsymbol{1}\in \mathbb{R}^{n}$ and a covariance matrix $\boldsymbol{\Sigma}_{k} = \sigma^{2}_{k}(\boldsymbol{x})\boldsymbol{I} \in \mathbb{S}^{n}_{++}$. The number of mixture components, $C$, equals the number of classes in a dataset. The parameters $\pi_{k}$, $\boldsymbol{\mu}_{k}$, and $\boldsymbol{\Sigma}_{k}$ describe the presence, spatial positioning, and statistical complexity (measures of spread and uncertainty) of the $k$th class membership within the GMM, given the input sample $\boldsymbol{x}$. Following a common configuration technique \cite{chen2020simple, khosla2020supervised}, we instantiate the MDN as a multi-layer perceptron with two hidden layers (2048 and 128 nodes) and an output layer of size 3C, where each neuron corresponds to the GMM parameters. To obtain these parameters, we employ forward propagation on the MDN with appropriate activation functions, ensuring the constraints of the parameters are met. The softmax activation normalizes mixture coefficients, promoting positivity and a sum of 1, while the exponential linear unit (ELU) \cite{varamesh2020mixture} enforces semi-positivity in standard deviations. The detailed architecture of MDN can be found in Figure \ref{fig:probmcl} (c). We remove the MDN after contrastive training, resulting in inference-time models with the same parameters and memory footprint as the encoder \textit{Enc}($\cdot$).
\subsection{Multi-label Contrastive Learning with ProbMCL} \label{subsec:learning}
We now delve into the learning process for multi-label classification tasks within the \texttt{ProbMCL} framework. To achieve this, we propose an objective function comprising two terms: \textit{negative log-likelihood loss} and \textit{probabilistic contrastive loss}. For a set of $N$ randomly sampled image/label pairs, denoted as $\{\boldsymbol{x}_{i}, \boldsymbol{y}_{i}\}_{i=1}^{N}$, the corresponding training batch comprises $2N$ pairs, $\{\hat{\boldsymbol{x}}_{i}, \hat{\boldsymbol{y}}_{i}\}_{i=1}^{2N}$, where $\hat{\boldsymbol{x}}_{2i}$ and $\hat{\boldsymbol{x}}_{2i-1}$ are two random augmentations of $\boldsymbol{x}_{i}$ ($i= 1, \cdots, N$) and $\hat{\boldsymbol{y}}_{2i} = \hat{\boldsymbol{y}}_{2i-1} = \boldsymbol{y}_{i}$. Throughout this paper, $I:= \{1, \cdots, 2N\}$ and $C$ denote the mini-batch and total number of classes, respectively.

\textbf{Negative Log-likelihood Loss.} 
The parameters of the Gaussian mixture model defined in Eq \ref{eq:gmm} are obtained by optimizing the following negative log-likelihood loss with respect to the weights of the encoder network and MDN ($\boldsymbol{\theta}$) through using SGD:
\begin{flalign} \label{eq:nll}
    \mathcal{L}_{\text{NLL}}(\boldsymbol{\theta}) = \sum_{i\in I} -\log p(\boldsymbol{z}|\boldsymbol{x}_{i}, \boldsymbol{\theta}),
\end{flalign}
where $p(\boldsymbol{z}|\boldsymbol{x}_{i}, \boldsymbol{\theta})$ represents the Gaussian mixture density associated with the output feature of the $i$th sample parametrized by $\boldsymbol{\theta}$. Employing a negative log-likelihood loss function for a Gaussian mixture can result in numerical instabilities due to singularities and high data dimensions. These issues stem from mixture components with negligible contributions, which can lead to extremely small variances, resulting in irregular gradients that hinder proper model training \cite{varamesh2020mixture}. To mitigate these challenges, we adopt a modified ELU activation function for the variances, ensuring a minimum value of one.

\textbf{Probabilistic Contrastive Loss.} 
In multi-label representation learning, capturing correlations between frequently co-occurring semantic labels is crucial. To achieve this, we introduce a novel probabilistic contrastive loss, $\mathcal{L}_{\text{PCL}}$, which integrates label correlation and encoder uncertainty into contrastive learning. Unlike deterministic contrastive losses \cite{khosla2020supervised, zhang2022use}, $\mathcal{L}_{\text{PCL}}$ considers the uncertainty of the embedded distribution when handling positive and negative pairs. Its primary objective is to pull together the feature embeddings of an anchor image and its positive images in the probability space while simultaneously pushing apart negative samples lacking sufficient shared classes with the anchor. This is determined using an overlap decision threshold denoted as $\alpha \in [0, 1]$. Specifically, $A(i) = \{j \in I \backslash \{i\}: \mathcal{D}(\boldsymbol{y}_{i}, \boldsymbol{y}_{j})\geq \alpha\}$ defines the set of indices representing positive samples related to anchor $i$, where $\mathcal{D}(\cdot, \cdot)$ measures the overlap between two label vectors (e.g., Jaccard index or cosine similarity). The contrastive loss $\mathcal{L}_{\text{PCL}}$ for the entire minibatch is then formulated as follows:
\begin{flalign} \label{eq:pcl}
     \scalemath{0.95} {\sum_{i \in I} \dfrac{-1}{|A(i)|} \sum_{j\in A(i)} \mathcal{D}(\boldsymbol{y}_{i}, \boldsymbol{y}_{j}) \Big(\log\dfrac{\exp{\big(\text{Sim}(p_{i}, p_{j})/\tau\big)}}{\sum\limits_{l\in I\backslash \{i\}}\exp{\big(\text{Sim}(p_{i}, p_{l})/\tau\big)}} \Big)},
\end{flalign}
where $\text{Sim}(p_{i}, p_{j}) = \text{Sim}\big(p(\boldsymbol{z}|\boldsymbol{x}_{i}, \boldsymbol{\theta}), p(\boldsymbol{z}|\boldsymbol{x}_{j}, \boldsymbol{\theta})\big)$ is a similarity metric between two Gaussian mixtures like the Bhattacharyya coefficient \cite{bhattacharyya1946measure}, correlation coefficient \cite{sajedi2023new, kampa2011closed}, etc. The parameter $\tau$ represents the temperature, and $|\cdot|$ is the cardinality function. In this study, we adopt the correlation coefficient, given by $\text{Sim}(p_{i}, p_{j}) = \frac{\int_{\mathbb{R}^{n}} p(\boldsymbol{z}|\boldsymbol{x}_{i}, \boldsymbol{\theta})p(\boldsymbol{z}|\boldsymbol{x}_{j}, \boldsymbol{\theta})d{\boldsymbol{z}}}{\sqrt{\int_{\mathbb{R}^{n}} p(\boldsymbol{z}|\boldsymbol{x}_{i}, \boldsymbol{\theta})^{2}d{\boldsymbol{z}}}\sqrt{\int_{\mathbb{R}^{n}} p(\boldsymbol{z}|\boldsymbol{x}_{j}, \boldsymbol{\theta})^{2}d{\boldsymbol{z}}}}$, due to its closed-form solution for Gaussian mixtures \cite{sajedi2023new}. The overlap index $\mathcal{D}(\cdot, \cdot) \in [0, 1]$ acts as a weight for positive samples, emphasizing those sharing more labels with the anchor while devaluing samples with fewer shared labels. In the experimental section, we evaluate the effect of loss hyperparameters and the choice of $\mathcal{D}$ on \texttt{ProbMCL}'s performance.

\textbf{Objective Loss Function.}
The overall training loss for representation learning is formulated as the augmented Lagrangian of the two above-mentioned losses:
\begin{flalign} \label{eq:loss}
 \mathcal{L} = \mathcal{L}_{\text{NLL}} + \lambda \mathcal{L}_{\text{PCL}},
\end{flalign}
where $\lambda$ is a Lagrangian multiplier utilized to equalize the gradients of $\mathcal{L}_\text{NLL}$ and $\mathcal{L}_\text{PCL}$. During inference, the MDN is removed, and the original neural network (Enc($\cdot$)) is retained after training. To apply the trained model for multi-label classification, a linear classifier is then trained on the frozen representations, employing an asymmetric loss function $\mathcal{L}_{\text{ASL}}$ as described in \cite{ridnik2021asymmetric}.

\section{Experimental Results} \label{sec:experiments}

\subsection{Experimental Setup}  \label{subsec:setup}
\textbf{Datasets and Encoder Networks.}
We evaluate the performance of \texttt{ProbMCL} using different encoder networks on two benchmark multi-label image recognition datasets: MS-COCO \cite{lin2014microsoft}, a computer vision dataset, and ADP (Atlas of Digital Pathology) \cite{hosseini2019atlas, 9679989_samir}, a medical imaging dataset. We employ TResNet-M and TResNet-L \cite{ridnik2021asymmetric, ridnik2021tresnet} as state-of-the-art backbone encoders, derived from ResNet50 and ResNet101, respectively. These encoders are optimized for image resolutions of 224 and 448, respectively.

\textbf{Evaluation.} Consistent with established practices \cite{chen2019multi, zhu2021residual, sajedi2023end, zhao2021transformer, ridnik2021asymmetric}, we report the standard evaluation metrics, including mean average precision (mAP), average per-class precision (CP), recall (CR), and F1 score (CF1), alongside average overall precision (OP), recall (OR), and F1 score (OF1). We also assess computational costs using the number of parameters (M) and GMAC. Our evaluation employs an inference threshold of 0.5, categorizing labels as positive when confidence scores exceed this value \cite{chen2019multi, zhu2021residual}.

\textbf{Implementation Details.} We employed the \texttt{ProbMCL} framework in PyTorch, utilizing RandAugment \cite{sajedi2023end, wang2020multi} for augmentation in both the contrastive and classification stages. MDN parameters were uniformly initialized for mixture coefficients and mean vector weights, while covariance weights were set constant at 1. Training employed an Adam optimizer \cite{kingma2014adam} with an initial learning rate of 1e-4. The OneCycleLR scheduler \cite{smith2019super} ran for 80 epochs in contrastive and 40 epochs in classification, using a batch size of 128. Hyperparameters $\tau$ and $\lambda$ were tuned to 0.2 and 0.3, while $\alpha$ was set to 0.6 for MS-COCO and 0.5 for ADP (see Fig. \ref{fig:ablation}). Default asymmetric loss hyperparameters \cite{ridnik2021asymmetric} were used for classification. Experiments were conducted on dual NVIDIA GeForce RTX 2080Ti GPUs.

\subsection{Experimental Results and Discussion}

\begin{figure*}
    \centering
    \begin{subfigure}{0.33\textwidth}
        \includegraphics[width=\linewidth]{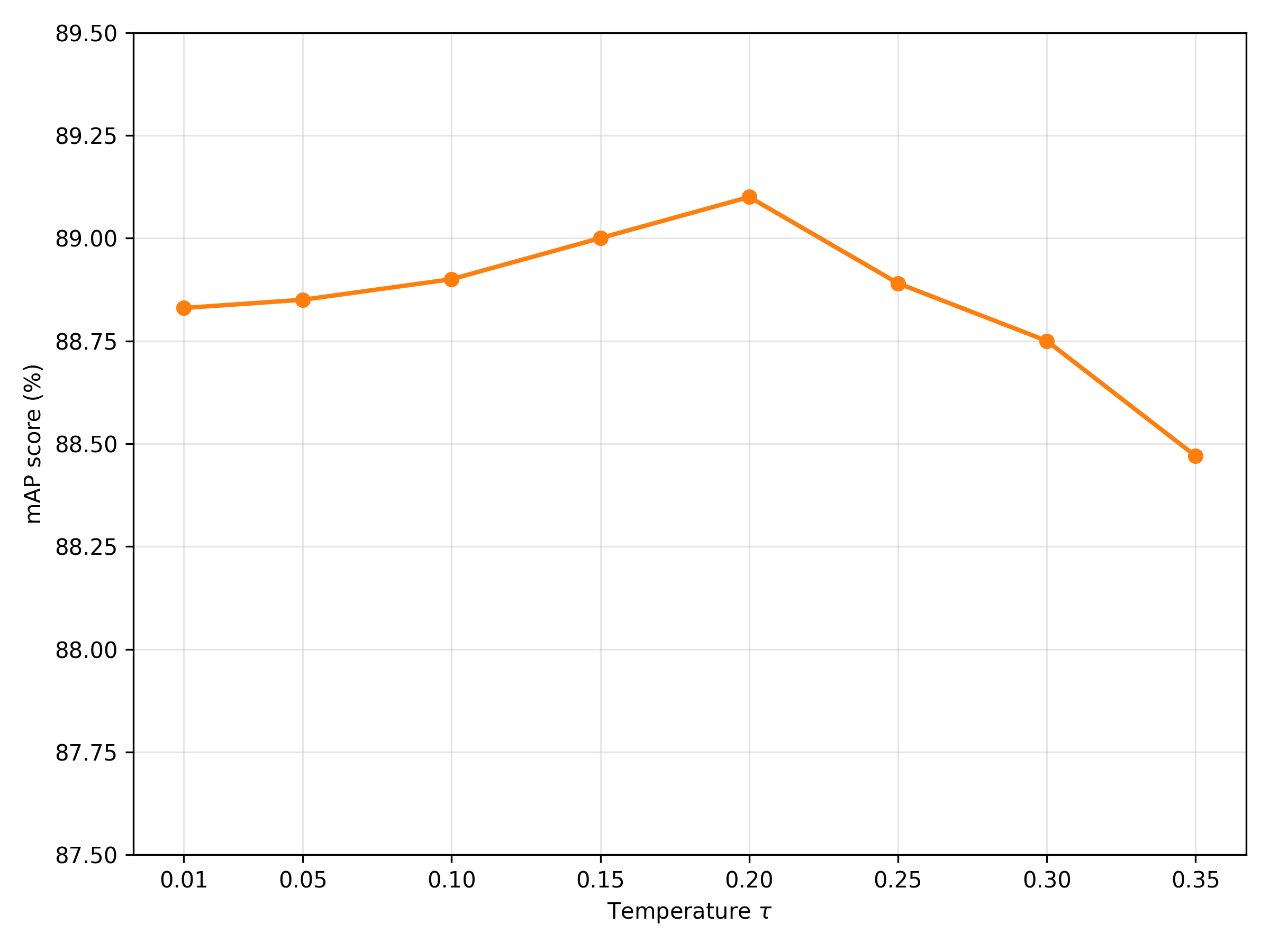}
        \caption{Ablation of temperature $\tau$.}
        \label{fig:temp}
    \end{subfigure}
    \begin{subfigure}{0.33\textwidth}
        \includegraphics[width=\linewidth]{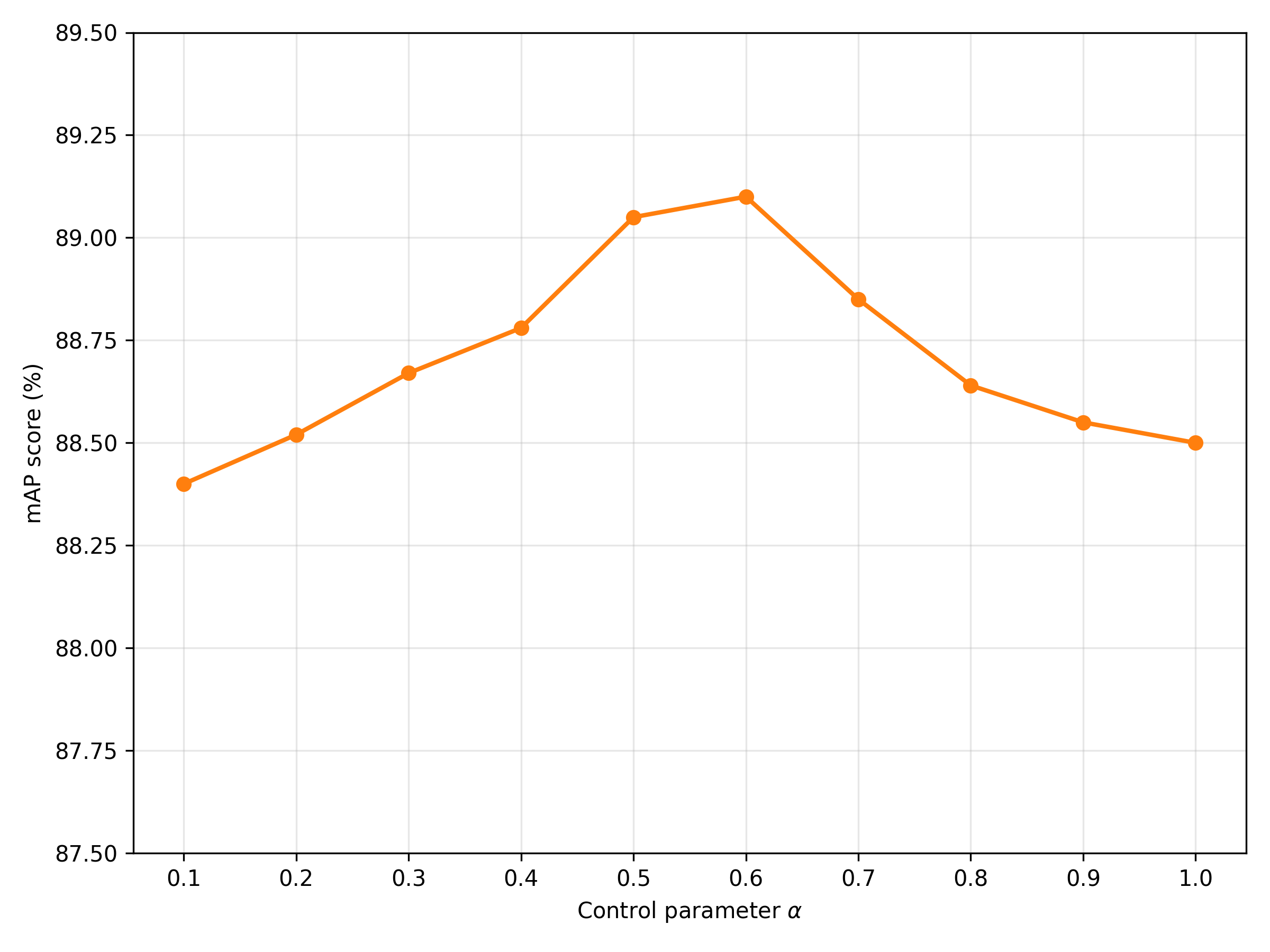}
        \caption{Ablation of control parameter $\alpha$.}
        \label{fig:alpha}
    \end{subfigure}
    \begin{subfigure}{0.33\textwidth}
        \includegraphics[width=\linewidth]{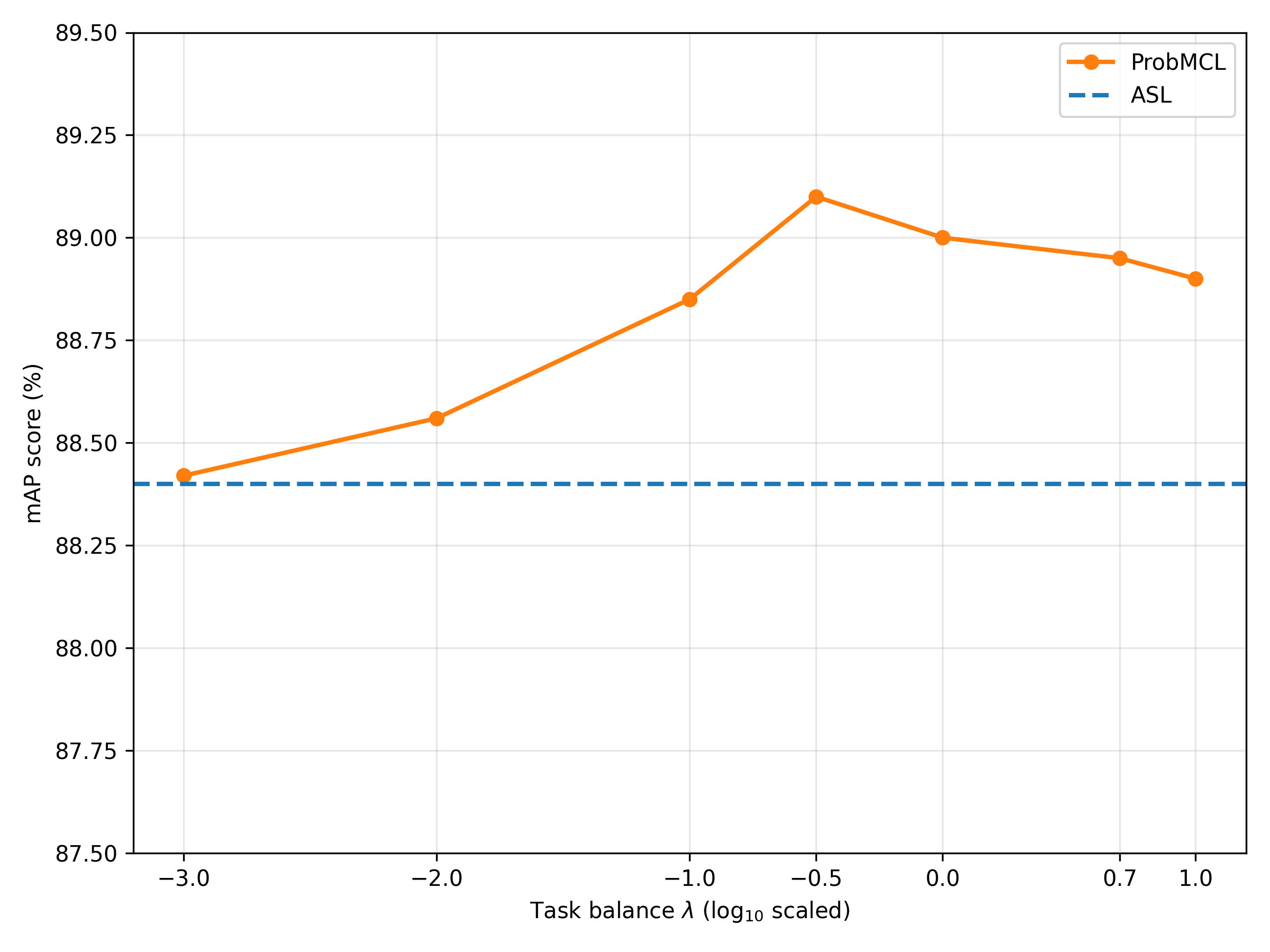}
        \caption{Ablation of task balance $\lambda$.}
        \label{fig:lambda}
    \end{subfigure}
\vspace{-10pt}
\caption{The effect of loss hyperparameters on the mAP score (\%) for the MS-COCO dataset \cite{lin2014microsoft}.}
\label{fig:ablation}
\end{figure*}
\begin{table} 
	\caption{Comparisons with prior methods on the MS-COCO dataset \cite{lin2014microsoft}. The upper and lower blocks correspond to ResNet-50 and ResNet-101 based models with input image resolutions of 224 and 448, respectively. \textbf{Bold entries} are the best results.}
	\label{tab:mscoco}
	\centering
	\footnotesize	
		\begin{tabular}{c|c|c|c|c|c|c|l}
			\hline
			Methods      & mAP        & CP  &  CR   & CF1  &  OP  &   OR   & OF1           \\ \hline\hline
			SRN~\cite{zhu2017learning}   &   77.1    &     81.6  &  65.4  &  71.2  &  82.7  &  69.9  & 75.8     \\
			PLA~\cite{yazici2020orderless}      & -    & 80.4  &  68.9  &  74.2  &  81.5  &  73.3  &  77.2 \\
			KMCL~\cite{sajedi2023end}      &   82.1   &  84.1  &  72.0  &  77.6  &  85.0  &  76.1  &  80.1     \\
			\textbf{ProbMCL} &  \bf{82.8}  &  \bf{85.0}  &  \bf{72.5}  & \bf{78.2}  &  \bf{85.3} &  \bf{76.9} &  \bf{80.9} \\
			\hline \hline
			ML-GCN~\cite{chen2019multi}     &   83.0  &  85.1  &  72.0  &  78.0  &  85.8  &  75.4  &  80.3         \\
			 KSSNet~\cite{wang2020multi}   & 83.7  &  84.6  &  73.2  &  77.2  &  \bf{87.8}  &  76.2  &  81.5 \\
			 TDRG~\cite{zhao2021transformer}    &  84.6  &  86.0  &  73.1  &  79.0  &  86.6  &  76.4  &  81.2 \\
             CSRA~\cite{zhu2021residual}   &  83.5  &  84.1  &  72.5  &  77.9  &  85.6  &  75.7  &  80.3 \\
             KMCL~\cite{sajedi2023end}    &  88.6  &  87.7  &  81.6  &  84.5 
            &  86.3  &  83.6  &  84.9\\
             ASL~\cite{ridnik2021asymmetric}   &  88.4  &  85.0  &  \bf{81.9}  &  83.4  &  85.2  &  84.2  &  84.7\\
             \textbf{ProbMCL} & \bf{89.1} & \bf{88.5} & \bf{81.9} & \bf{85.1} & 86.7 & \bf{84.3} & \bf{85.5} \\
			\hline
		\end{tabular}
\end{table}
\textbf{Performance Comparison with Prior Methods.} We evaluate the performance of \texttt{ProbMCL} against leading multi-label classification methods, including ML-GCN \cite{chen2019multi}, TDRG \cite{zhao2021transformer}, CSRA \cite{zhu2021residual}, ASL \cite{ridnik2021asymmetric}, and KMCL \cite{sajedi2023end}, on computer vision and medical imaging datasets. Table \ref{tab:mscoco} illustrates \texttt{ProbMCL}'s superiority over its competitors on MS-COCO, excelling across various metrics. Leveraging the TResNet-M encoder at a resolution of 224 achieves state-of-the-art results with a $0.7\%$ higher mAP than the best alternative. Similarly, TResNet-L at 448 resolution highlights \texttt{ProbMCL}'s dominance in overall and per-class metrics. The evaluation extends to medical imaging (see Table \ref{tab:asl}), where \texttt{ProbMCL} balances precision and recall, notably enhancing recall while maintaining competitive precision, including a leading mAP. This is significant, as recall indicates the risk of missing medical diagnoses. These achievements stem from generating valid positive examples using contrastive learning and improving representations to capture label co-dependencies. \texttt{ProbMCL} exposes the model to meaningful outcomes beyond ground truth, thus improving prediction accuracy through enhanced classification loss. Furthermore, our method shows a reduced computational footprint compared to previous approaches utilizing heavy-duty modules, as evidenced in Table \ref{tab:cost}. These findings highlight \texttt{ProbMCL}'s benefit for training in computationally limited environments.\\
\begin{table}
	\caption{Comparisons with state-of-the-art methods on the ADP dataset \cite{hosseini2019atlas}. \textbf{Bold entries} are the best results.}
	\label{tab:asl}
	\centering
	\footnotesize	
		\begin{tabular}{c|c|c|c|c|c|c|l}
			\hline
			Methods      & mAP        & CP  &  CR   & CF1  &  OP  &   OR   & OF1     \\ \hline\hline
			ML-GCN~\cite{chen2019multi}     &   94.9  &  91.8  &  87.0  &  89.3  &  92.0  &  86.9  &  89.7\\
			 TDRG~\cite{zhao2021transformer}    &  95.5  &  \bf{94.6}  &  84.8  &  89.4  &  \bf{94.3}  &  86.2  &  90.5 \\
             CSRA~\cite{zhu2021residual}   &  96.1  &  93.1  &  88.6  &  90.8  &  93.0  &  89.7  &  91.7 \\
             KMCL~\cite{sajedi2023end}    &  96.5  &  92.6  &  92.0  &  92.3  &  92.7  &  92.9  &  92.8\\
             ASL~\cite{ridnik2021asymmetric}   &  96.1  &  93.1  &  88.6  &  90.8  &  92.1  &  90.7  &  91.4\\
             \textbf{ProbMCL} & \bf{96.9}  & 93.0 & \bf{92.7} & \bf{92.8} & 92.9 & \bf{93.3} & \bf{93.1} \\
			\hline
		\end{tabular}
\end{table}
\begin{table} 
	\caption{Computational training cost comparison with prior methods on the MS-COCO dataset \cite{lin2014microsoft}. \textbf{Bold entries} are the best results.}
	\label{tab:cost}
	\centering
	\footnotesize	
		\begin{tabular}{c|c|c|c|c}
			\hline
			Methods    & ML-GCN\cite{chen2019multi}        & TDRG\cite{zhao2021transformer}  &  CSRA\cite{zhu2021residual}  &  \textbf{ProbMCL}      \\ \hline\hline
			Parameters   &  44.90 M  & 75.20 M  &  42.52 M &   \bf{42.23} M \\
			 GMAC   & 31.39  &  64.40  & 31.39 &  \bf{29.65}  \\
			\hline
		\end{tabular}
\end{table}
\textbf{Ablation Study on Overlapping Index Function.} In this section, we examine the impact of different overlapping functions $\mathcal{D}(\cdot, \cdot)$ on positive set construction and \texttt{ProbMCL}'s effectiveness using the MS-COCO dataset. We employ the Jaccard index $J(\mathbf{y}_{i}, \mathbf{y}_{j}) = \frac{\mathbf{y}_{i} \boldsymbol{\cdot} \mathbf{y}_{j}}{\|\mathbf{y}_{i}\|^{2}+\|\mathbf{y}_{j}\|^{2}-\mathbf{y}_{i}\boldsymbol{\cdot}\mathbf{y}_{j}}$ and cosine similarity $\cos(\mathbf{y}_{i}, \mathbf{y}_{j}) = \frac{\mathbf{y}_{i} \boldsymbol{\cdot} \mathbf{y}_{j}}{\|\mathbf{y}_{i}\|^{2} \|\mathbf{y}_{j}\|^{2}}$ as $\mathcal{D}(\mathbf{y}_{i}, \mathbf{y}_{j})$ in Eq. \ref{eq:pcl} during contrastive learning. Experimental results show the Jaccard index's superiority (89.1\% mAP) due to consideration of object co-occurrences and label overlap via Intersection over Union (IoU). In contrast, the cosine similarity (88.9\% mAP) only measures the angle between label vectors. Thus, the Jaccard index is preferred for set construction.

\textbf{Ablation Study on Loss Hyperparameters.} We conduct ablation experiments on temperature ($\tau$), control parameter ($\alpha$), and task balance ($\lambda$) for MS-COCO classification. Fig. \ref{fig:temp} demonstrates that a lower temperature $\tau$ enhances training, yet excessive lowering leads to numerical instability. We use $\tau =$ 0.2 for better learning from hard negative samples. The control parameter $\alpha$ significantly impacts performance. In Fig. \ref{fig:alpha}, smaller $\alpha$ values worsen performance compared to higher values. A smaller $\alpha$ enlarges the positive set, potentially overlooking label co-dependencies, while a larger $\alpha$ shrinks the positive set and can miss partial label correlations. The optimal results are at $\alpha$ = 0.6. Finally, we evaluate the impact of perturbing the contribution of $\mathcal{L}_{\text{NLL}}$ and $\mathcal{L}_{\text{PCL}}$ on contrastive learning. Lower $\lambda$ reduces the contrastive effect (resembling ASL), while higher values enhance the contrastive behavior and improve performance (Fig. \ref{fig:lambda}). Results show partial robustness to increased regularization at plateaus above 0.3, so we chose 0.3 for experiments.
 \begin{figure} [H]
    \centering
    \vspace{-11pt}
    \includegraphics[width= 1.0\linewidth]{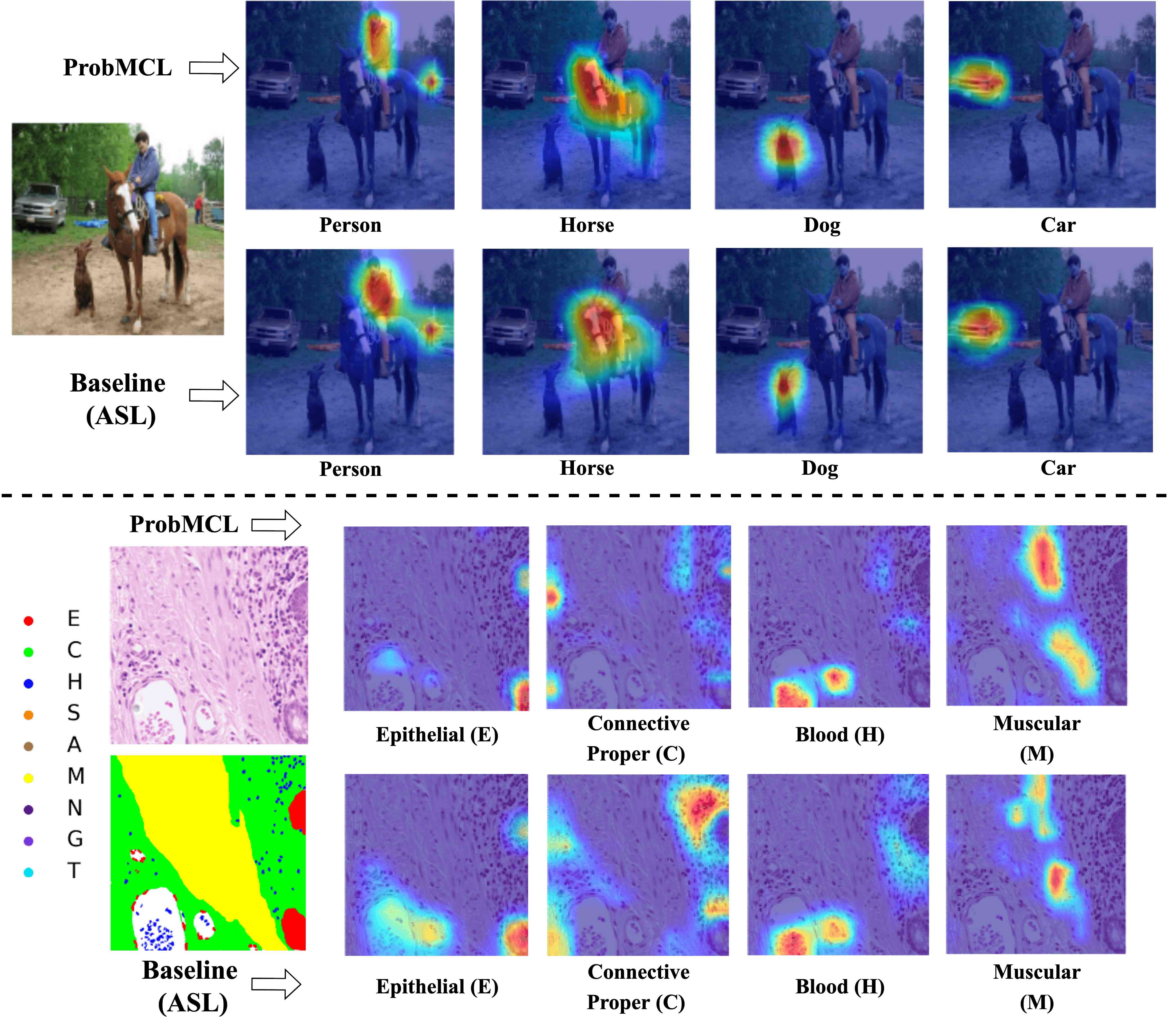}
    \caption{Visualization analyses of baseline (ASL) and the proposed method across MS-COCO (top) and ADP (bottom) datasets.}
    \label{fig:gradcam}
\end{figure}

\vspace{-5pt}
\textbf{Visualizations.} We utilized Grad-CAM \cite{selvaraju2017grad} to depict the visualization results of both the baseline (ASL) and the proposed \texttt{ProbMCL} in Fig. \ref{fig:gradcam}. These visualizations highlight our approach's superiority in contrasting dissimilar-looking objects; e.g., \texttt{ProbMCL} effectively distinguishes the person and horse labels with minimal extraneous activations. Moreover, by capturing label correlation and uncertainty from the novel representation, \texttt{ProbMCL} accurately identifies localization and precisely perceives small objects, as seen with the Person and Blood (H) classes in Fig. \ref{fig:gradcam}.
\section{Conclusion} \label{conclusion}
In this work, we propose \texttt{ProbMCL}, a simple yet effective framework for multi-label image classification. The framework integrates multi-label contrastive learning with Gaussian mixture latent space, aimed at capturing label dependencies and exploring encoder uncertainty to enhance model performance. The quantitative and qualitative results confirm the superiority of \texttt{ProbMCL} over existing techniques with low computational training costs across computer vision and medical imaging datasets. In the future, we plan to extend our approach to include segmentation and detection tasks and explore its applicability to modalities like natural language processing.


\newcommand{\BIBdecl}{\setlength{\itemsep}{0em}}
\bibliographystyle{IEEEtran}
\bibliography{main}

\end{document}